\documentclass[11pt,a4paper]{article}
\usepackage[hyperref]{emnlp-ijcnlp-2019}
\usepackage{times}
\usepackage{latexsym}
\usepackage{amsmath}
\usepackage{amssymb}
\usepackage{url}
\usepackage{textcomp}
\usepackage{graphicx}
\usepackage{placeins}

\aclfinalcopy 

\title{Sentence-BERT: Sentence Embeddings using Siamese BERT-Networks}

\author{Nils Reimers and Iryna Gurevych \\
Ubiquitous Knowledge Processing Lab (UKP-TUDA)\\
Department of Computer Science, Technische Universit\"at Darmstadt\\
\url{www.ukp.tu-darmstadt.de}}

\date{}

\begin{document}
\maketitle
\begin{abstract}
BERT \cite{devlin2018bert} and RoBERTa  \cite{roberta} has set a new state-of-the-art performance on sentence-pair regression tasks like semantic textual similarity (STS). However, it requires that both sentences are fed into the network, which causes a massive computational overhead: Finding the most similar pair in a collection of 10,000 sentences requires about 50 million inference computations (\texttildelow 65 hours) with BERT. The construction of BERT makes it unsuitable for semantic similarity search as well as for unsupervised tasks like clustering. 

In this publication, we present Sentence-BERT (SBERT), a modification of the pretrained BERT network that use siamese and triplet network structures to derive semantically meaningful sentence embeddings that can be compared using cosine-similarity. This reduces the effort for finding the most similar pair from 65 hours with BERT / RoBERTa to about 5 seconds with SBERT, while maintaining the accuracy from BERT.

We evaluate SBERT and SRoBERTa on common STS tasks and transfer learning tasks, where it outperforms other state-of-the-art sentence embeddings methods.\footnote{Code available: \url{https://github.com/UKPLab/sentence-transformers}}
\end{abstract}

\section{Introduction}
In this publication, we present Sentence-BERT (SBERT), a modification of the BERT network using siamese and triplet networks that is able to derive semantically meaningful sentence embeddings\footnote{With \textit{semantically meaningful} we mean that semantically similar sentences are close in vector space.}. This enables BERT to be used for certain new tasks, which up-to-now were not applicable for BERT. These tasks include large-scale semantic similarity comparison, clustering, and information retrieval via semantic search.

BERT set new state-of-the-art performance on various sentence classification and sentence-pair regression tasks. BERT uses a cross-encoder: Two sentences are passed to the transformer network and the target value is predicted. However, this setup is unsuitable for various pair regression tasks due to too many possible combinations. Finding in a collection of $n=10\,000$ sentences the pair with the highest similarity requires with BERT $n\cdot(n-1)/2=49\,995\,000$ inference computations. On a modern  V100 GPU, this requires about 65 hours. Similar, finding which of the over 40 million existent questions of Quora is the most similar for a new question could be modeled as a pair-wise comparison with BERT, however, answering a single query would require over 50 hours.

A common method to address clustering and semantic search is to map each sentence to a vector space such that semantically similar sentences are close. Researchers have started to input individual sentences into BERT and to derive fixed-size sentence embeddings. The most commonly used approach is to average the BERT output layer (known as BERT embeddings) or by using the output of the first token (the \texttt{[CLS]} token). As we will show, this common practice yields rather bad sentence embeddings, often worse than averaging GloVe embeddings \cite{glove}.

To alleviate this issue, we developed SBERT. The siamese network architecture enables that fixed-sized vectors for input sentences can be derived. Using a similarity measure like cosine-similarity or Manhatten / Euclidean distance, semantically similar sentences can be found. These similarity measures can be performed extremely efficient on modern hardware, allowing SBERT to be used for semantic similarity search as well as for clustering. The complexity for finding the most similar sentence pair in a collection of 10,000 sentences is reduced from 65 hours with BERT to the computation of 10,000 sentence embeddings (\texttildelow 5 seconds with SBERT) and computing cosine-similarity (\texttildelow 0.01 seconds). By using optimized index structures, finding the most similar Quora question can be reduced from 50 hours to a few milliseconds \cite{JDH17}.

We fine-tune SBERT on NLI data, which creates sentence embeddings that significantly outperform other state-of-the-art sentence embedding methods like InferSent \cite{conneau2017infersent} and Universal Sentence Encoder \cite{universal_sentence_encoder}. On seven Semantic Textual Similarity (STS) tasks, SBERT achieves an improvement of 11.7 points compared to InferSent and 5.5 points compared to Universal Sentence Encoder. On SentEval \cite{conneau2018senteval}, an evaluation toolkit for sentence embeddings, we achieve an improvement of 2.1 and 2.6 points, respectively.

SBERT can be adapted to a specific task. It sets new state-of-the-art performance on a challenging argument similarity dataset \cite{MisraEW16} and on a triplet dataset to distinguish sentences from different sections of a Wikipedia article \cite{ein-dor-etal-2018-learning}.

The paper is structured in the following way: Section \ref{sec_model} presents SBERT, section \ref{sec_eval_sts} evaluates SBERT on common STS tasks and on the challenging Argument Facet Similarity (AFS) corpus \cite{MisraEW16}. Section \ref{sec_eval_senteval} evaluates SBERT on SentEval. In section \ref{sec_ablation_study}, we perform an ablation study to test some design aspect of SBERT. In section \ref{sec_computational_efficiency}, we compare the computational efficiency of SBERT sentence embeddings in contrast to other state-of-the-art sentence embedding methods.

\section{Related Work}
We first introduce BERT, then, we discuss state-of-the-art sentence embedding methods.

BERT \cite{devlin2018bert} is a pre-trained transformer network \cite{Attention_is_all_you_need}, which set for various NLP tasks new state-of-the-art results, including question answering, sentence classification, and sentence-pair regression. The input for BERT for sentence-pair regression consists of the two sentences, separated by a special \texttt{[SEP]} token. Multi-head attention over 12 (base-model) or 24 layers (large-model) is applied and the output is passed to a simple regression function to derive the final label. Using this setup, BERT set a new state-of-the-art performance on the Semantic Textual Semilarity (STS) benchmark \cite{sts2017}. RoBERTa \cite{roberta} showed, that the performance of BERT can further improved by small adaptations to the pre-training process. We also tested XLNet \cite{xlnet}, but it led in general to worse results than BERT. 

A large disadvantage of the BERT network structure is that no independent sentence embeddings are computed, which makes it difficult to derive sentence embeddings from BERT. To bypass this limitations, researchers passed single sentences through BERT and then derive a fixed sized vector by either averaging the outputs (similar to average word embeddings) or by using the output of the special \texttt{CLS} token (for example: \newcite{bert_sentence_embeddings_1,bert_sentence_embeddings_2,bert_sentence_embeddings_3}). These two options are also provided by the popular bert-as-a-service-repository\footnote{\url{https://github.com/hanxiao/bert-as-service/}}. Up to our knowledge, there is so far no evaluation if these methods lead to useful sentence embeddings.

Sentence embeddings are a well studied area with dozens of proposed methods. Skip-Thought \cite{SkipThought} trains an encoder-decoder architecture to predict the surrounding sentences. InferSent \cite{conneau2017infersent} uses labeled data of the Stanford Natural Language Inference dataset \cite{snli} and the Multi-Genre NLI dataset \cite{multinli} to train a siamese BiLSTM network with max-pooling over the output. Conneau et al.\ showed, that InferSent consistently outperforms unsupervised methods like SkipThought.  Universal Sentence Encoder \cite{universal_sentence_encoder} trains a transformer network and augments unsupervised learning with training on SNLI. \newcite{hill-etal-2016-learning} showed, that the task on which sentence embeddings are trained significantly impacts their quality.  Previous work \cite{conneau2017infersent, universal_sentence_encoder}  found that the SNLI datasets are suitable for training sentence embeddings. \newcite{yang-2018-learning} presented a method to train on conversations from Reddit using siamese DAN and siamese transformer networks, which yielded good results on the STS benchmark dataset.

\newcite{polyencoders} addresses the run-time overhead of the cross-encoder from BERT and present a method (poly-encoders) to compute a score between $m$ context vectors and pre-computed candidate embeddings using attention. This idea works for finding the highest scoring sentence in a larger collection. However, poly-encoders have the drawback that the score function is not symmetric and the computational overhead is too large for use-cases like clustering, which would require $O(n^2)$ score computations. 

Previous neural sentence embedding methods started the training from a random initialization. In this publication, we use the pre-trained BERT and RoBERTa network and only fine-tune it to yield useful sentence embeddings. This reduces significantly the needed training time: SBERT can be tuned in less than 20 minutes, while yielding better results than comparable sentence embedding methods.

\section{Model} \label{sec_model}

\begin{figure}[t]
	\centering
	\includegraphics[width=0.8\linewidth]{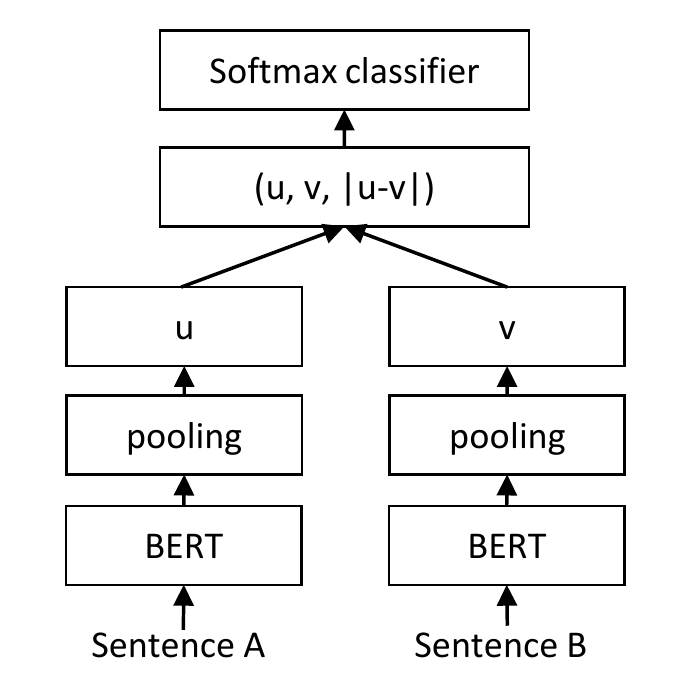}
	\caption{SBERT architecture with classification objective function, e.g., for fine-tuning on SNLI dataset. The two BERT networks have tied weights (siamese network structure).}
	\label{fig_sbert_softmax}
\end{figure}

SBERT adds a pooling operation to the output of BERT / RoBERTa to derive a fixed sized sentence embedding.  We experiment with three pooling strategies: Using the output of the \texttt{CLS}-token, computing the mean of all output vectors (\texttt{MEAN}-strategy), and computing a max-over-time of the output vectors (\texttt{MAX}-strategy). The default configuration is \texttt{MEAN}.

In order to fine-tune BERT / RoBERTa, we create siamese and triplet networks \cite{triplet_network} to update the weights such that the produced sentence embeddings are semantically meaningful and can be compared with cosine-similarity.

The network structure depends on the available training data. We experiment with the following structures and objective functions.

\textbf{Classification Objective Function.} We concatenate the sentence embeddings $u$ and $v$ with the element-wise difference $|u-v|$ and multiply it with the trainable weight $W_t \in \mathbb{R}^{3n \times k}$:
$$o = \text{softmax}(W_t (u, v, |u-v|))$$

where $n$ is the dimension of the sentence embeddings and $k$ the number of labels. We optimize cross-entropy loss. This structure is depicted in Figure \ref{fig_sbert_softmax}.

\begin{figure}[t]
	\centering
	\includegraphics[width=0.8\linewidth]{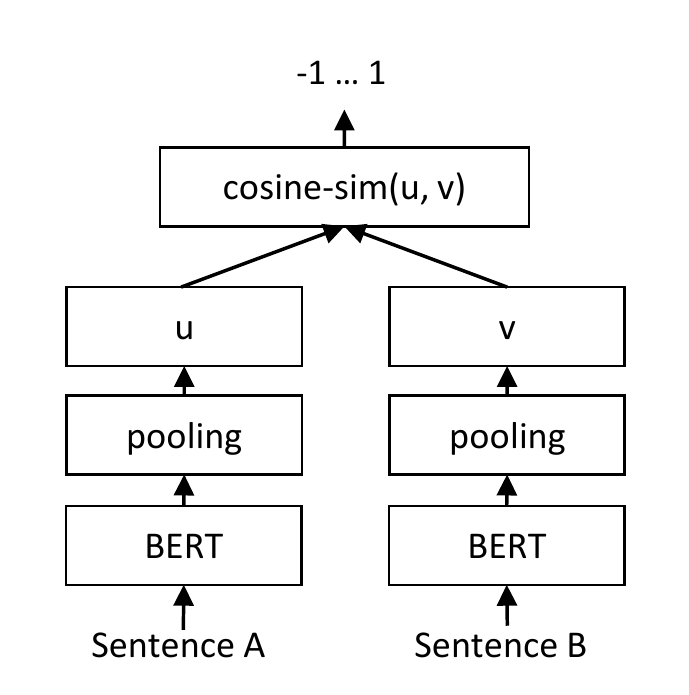}
	\caption{SBERT architecture at inference, for example, to compute similarity scores. This architecture is also used with the regression objective function.}
	\label{fig_sbert_cosine}
\end{figure}

\textbf{Regression Objective Function.} The cosine-similarity between the two sentence embeddings $u$ and $v$ is computed (Figure \ref{fig_sbert_cosine}). We use mean-squared-error loss as the objective function.

\textbf{Triplet Objective Function.} Given an anchor sentence $a$, a positive sentence $p$, and a negative sentence $n$, triplet loss tunes the network such that the distance between $a$ and $p$ is smaller than the distance between $a$ and $n$. Mathematically, we minimize the following loss function:
$$max(||s_a-s_p|| - ||s_a-s_n|| + \epsilon, 0)$$

with $s_x$ the sentence embedding for $a$/$n$/$p$, $||\cdot||$ a distance metric and margin $\epsilon$. Margin $\epsilon$ ensures that $s_p$ is at least $\epsilon$ closer to $s_a$ than $s_n$. As metric we use Euclidean distance and we set $\epsilon=1$ in our experiments.

\begin{table*}[t]
	\centering 
	\footnotesize
	\begin{tabular}{|l|c|c|c|c|c|c|c||c|}
		\hline
		\textbf{Model} & \textbf{STS12} & \textbf{STS13} & \textbf{STS14} & \textbf{STS15} & \textbf{STS16} & \textbf{STSb} & \textbf{SICK-R} & \textbf{Avg.} \\ \hline
		Avg.\ GloVe embeddings & 55.14 & 70.66 & 59.73 & 68.25 & 63.66 & 58.02 & 53.76 & 61.32 \\
		Avg.\ BERT embeddings & 38.78 & 57.98 & 57.98 & 63.15 & 61.06 & 46.35 & 58.40 & 54.81 \\
		BERT CLS-vector & 20.16 & 30.01 & 20.09 & 36.88 & 38.08 & 16.50 & 42.63 & 29.19 \\
		InferSent - Glove & 52.86 & 66.75 & 62.15 & 72.77 & 66.87 & 68.03 & 65.65 &  65.01 \\
		Universal Sentence Encoder & 64.49 & 67.80 & 64.61 & 76.83 & 73.18 & 74.92 & \textbf{76.69} & 71.22 \\ \hline
		SBERT-NLI-base & 70.97 & 76.53 & 73.19 & 79.09 & 74.30 & 77.03 & 72.91 & 74.89 \\
		SBERT-NLI-large & {72.27} & \textbf{78.46} & \textbf{74.90} & {80.99} & {76.25} & \textbf{79.23} & 73.75 & {76.55} \\ \hline 
		SRoBERTa-NLI-base & 71.54 & 72.49 & 70.80 & 78.74 & 73.69 & 77.77 & 74.46 & 74.21  \\ 
		SRoBERTa-NLI-large & \textbf{74.53} & 77.00 & 73.18 & \textbf{81.85} & \textbf{76.82} & 79.10 & 74.29 & \textbf{76.68} \\ \hline
	\end{tabular}
	\caption{Spearman rank correlation $\rho$ between the cosine similarity of sentence representations and the gold labels for various Textual Similarity (STS) tasks. Performance is reported by convention as $\rho \times 100$. STS12-STS16: SemEval 2012-2016, STSb: STSbenchmark, SICK-R: SICK relatedness dataset.}
	\label{table_sts_tasks}
\end{table*}

\subsection{Training Details}
We train SBERT on the combination of the SNLI \cite{snli} and the Multi-Genre NLI \cite{multinli} dataset. The SNLI is a collection of 570,000 sentence pairs annotated with the labels \textit{contradiction}, \textit{eintailment}, and \textit{neutral}. MultiNLI contains 430,000 sentence pairs and covers a range of genres of spoken and written text. We fine-tune SBERT with a 3-way softmax-classifier objective function for one epoch. We used a batch-size of 16, Adam optimizer with learning rate $2\mathrm{e}{-5}$, and a linear learning rate warm-up over 10\% of the training data. Our default pooling strategy is \texttt{MEAN}.

\section{Evaluation - Semantic Textual Similarity} \label{sec_eval_sts}

We evaluate the performance of SBERT for common Semantic Textual Similarity (STS) tasks. State-of-the-art methods often learn a (complex) regression function that maps sentence embeddings to a similarity score. However, these regression functions work pair-wise and due to the combinatorial explosion those are often not scalable if the collection of sentences reaches a certain size. Instead, we always use cosine-similarity to compare the similarity between two sentence embeddings. We ran our experiments also with negative Manhatten and negative Euclidean distances as similarity measures, but the results for all approaches remained roughly the same.

\subsection{Unsupervised STS}
We evaluate the performance of SBERT for STS without using any STS specific training data. We use the STS tasks 2012 - 2016 \cite{sts2012, sts2013, sts2014, sts2015, sts2016}, the STS benchmark \cite{sts2017}, and the SICK-Relatedness dataset \cite{sick}. These datasets provide labels between 0 and 5 on the semantic relatedness of sentence pairs. We showed in \cite{Reimers2016_STS} that Pearson correlation is badly suited for STS. Instead, we compute the Spearman's rank correlation between the cosine-similarity of the sentence embeddings and the gold labels. The setup for the other sentence embedding methods is equivalent, the similarity is computed by cosine-similarity. The results are depicted in Table \ref{table_sts_tasks}. 

The results shows that directly using the output of BERT leads to rather poor performances. Averaging the BERT embeddings achieves an average correlation of only 54.81, and using the \texttt{CLS}-token output only achieves an average correlation of 29.19. Both are worse than computing average GloVe embeddings. 

Using the described siamese network structure and fine-tuning mechanism substantially improves the correlation, outperforming both InferSent and Universal Sentence Encoder substantially. The only dataset where SBERT performs worse than Universal Sentence Encoder is SICK-R. Universal Sentence Encoder was trained on various datasets, including news, question-answer pages and discussion forums, which appears to be more suitable to the data of SICK-R. In contrast, SBERT was pre-trained only on Wikipedia (via BERT) and on NLI data.

While RoBERTa was able to improve the performance for several supervised tasks, we only observe minor difference between SBERT and SRoBERTa for generating sentence embeddings.

\subsection{Supervised STS} \label{sec_supervised_sts}

The STS benchmark (STSb) \cite{sts2017} provides is a popular dataset to evaluate supervised STS systems. The data includes 8,628 sentence pairs from the three categories \textit{captions}, \textit{news}, and \textit{forums}. It is divided into train (5,749), dev (1,500) and test (1,379). BERT set a new state-of-the-art performance on this dataset by passing both sentences to the network and using a simple regression method for the output.

\begin{table}[h]
	\centering 
	\footnotesize
	\begin{tabular}{|l|c|}
		\hline
		\textbf{Model} & \textbf{Spearman} \\ \hline
		\multicolumn{2}{|l|}{\textit{Not trained for STS}} \\ \hline
		Avg.\ GloVe embeddings & 58.02\\
		Avg.\ BERT embeddings &  46.35\\
		InferSent - GloVe &  68.03 \\
		Universal Sentence Encoder &  74.92\\
		SBERT-NLI-base  &  77.03\\
		SBERT-NLI-large & 79.23 \\ \hline 
		\multicolumn{2}{|l|}{\textit{Trained on STS benchmark dataset}} \\ \hline
		BERT-STSb-base & 84.30 $\pm$ 0.76  \\
		SBERT-STSb-base & 84.67 $\pm$ 0.19 \\ 
		SRoBERTa-STSb-base & \textbf{84.92} $\pm$ 0.34 \\ \hline 
		
		BERT-STSb-large  & \textbf{85.64} $\pm$ 0.81 \\ 
		SBERT-STSb-large & 84.45 $\pm$ 0.43 \\ 
		SRoBERTa-STSb-large & 85.02 $\pm$ 0.76 \\ \hline 
		
		\multicolumn{2}{|l|}{\textit{Trained on NLI data + STS benchmark data}} \\ \hline
		
		BERT-NLI-STSb-base & \textbf{88.33} $\pm$ 0.19 \\ 
		SBERT-NLI-STSb-base & 85.35 $\pm$ 0.17 \\ 
		SRoBERTa-NLI-STSb-base & 84.79 $\pm$ 0.38 \\ \hline
		
		BERT-NLI-STSb-large & \textbf{88.77} $\pm$ 0.46 \\ 
		SBERT-NLI-STSb-large & 86.10 $\pm$ 0.13  \\
		SRoBERTa-NLI-STSb-large & 86.15 $\pm$ 0.35 \\ \hline 
	\end{tabular}
	\caption{Evaluation on the STS benchmark test set. BERT systems were trained with 10 random seeds and 4 epochs. SBERT was fine-tuned on the STSb dataset, SBERT-NLI was pretrained on the NLI datasets, then fine-tuned on the STSb dataset. }
	\label{table_stsb}
\end{table}

We use the training set to fine-tune SBERT using the regression objective function. At prediction time, we compute the cosine-similarity between the sentence embeddings. All systems are trained with 10 random seeds to counter variances \cite{reimers_single_perf_score}.

The results are depicted in Table \ref{table_stsb}. We experimented with two setups: Only training on STSb, and first training on NLI, then training on STSb. We observe that the later strategy leads to a slight improvement of 1-2 points. This two-step approach had an especially large impact for the BERT cross-encoder, which improved the performance by 3-4 points. We do not observe a significant difference between BERT and RoBERTa.

\subsection{Argument Facet Similarity}
We evaluate SBERT on the Argument Facet Similarity (AFS) corpus by \newcite{MisraEW16}. The AFS corpus annotated 6,000 sentential argument pairs from social media dialogs on three controversial topics: \textit{gun control}, \textit{gay marriage}, and \textit{death penalty}. The data was annotated on a scale from 0 (``different topic") to 5 (``completely equivalent"). The similarity notion in the AFS corpus is fairly different to the similarity notion in the STS datasets from SemEval. STS data is usually descriptive, while AFS data are argumentative excerpts from dialogs. To be considered similar, arguments must not only make similar claims, but also provide a similar reasoning. Further, the lexical gap between the sentences in AFS is much larger. Hence, simple unsupervised methods as well as state-of-the-art STS systems perform badly on this dataset \cite{reimers-etal-2019-classification}.

We evaluate SBERT on this dataset in two scenarios: 1) As proposed by Misra et al., we evaluate SBERT using 10-fold cross-validation. A draw-back of this evaluation setup is that it is not clear how well approaches generalize to different topics. Hence, 2) we evaluate SBERT in a cross-topic setup. Two topics serve for training and the approach is evaluated on the left-out topic. We repeat this for all three topics and average the results.

SBERT is fine-tuned using the Regression Objective Function. The similarity score is computed using cosine-similarity based on the sentence embeddings. We also provide the Pearson correlation $r$ to make the results comparable to Misra et al. However, we showed \cite{Reimers2016_STS} that Pearson correlation has some serious drawbacks and should be avoided for comparing STS systems. The results are depicted in Table \ref{table_afs}. 

Unsupervised methods like tf-idf, average GloVe embeddings or InferSent perform rather badly on this dataset with low scores. Training SBERT in the 10-fold cross-validation setup gives a performance that is nearly on-par with BERT. 

However, in the cross-topic evaluation, we observe a performance drop of SBERT by about 7 points Spearman correlation. To be considered similar, arguments should address the same claims and provide the same reasoning. BERT is able to use attention to compare directly both sentences (e.g.\ word-by-word comparison), while SBERT must map individual sentences from an unseen topic to a vector space such that arguments with similar claims and reasons are close. This is a much more challenging task, which appears to require more than just two topics for training to work on-par with BERT.  

\begin{table}[h]
	\centering 
	\footnotesize
	\begin{tabular}{|l|c|c|}
		\hline
		\textbf{Model} & $r$ & $\rho$ \\ \hline
		\multicolumn{3}{|l|}{\textit{Unsupervised  methods}} \\ \hline
		tf-idf & 46.77 & 42.95 \\
		Avg.\ GloVe embeddings & 32.40 & 34.00 \\
		InferSent - GloVe & 27.08 &  26.63  \\
		\hline 
		\multicolumn{3}{|l|}{\textit{10-fold Cross-Validation}} \\ \hline
		SVR \cite{MisraEW16} & 63.33 & - \\
		BERT-AFS-base & {77.20}  & {74.84}  \\
		SBERT-AFS-base & 76.57 & 74.13 \\
		BERT-AFS-large & {78.68} & {76.38}  \\
		SBERT-AFS-large & 77.85 & 75.93 \\
		\hline
		\multicolumn{3}{|l|}{\textit{Cross-Topic Evaluation}} \\ \hline
		BERT-AFS-base & {58.49} & {57.23} \\ 
		SBERT-AFS-base & 52.34  & 50.65 \\ 
		BERT-AFS-large & {62.02} & {60.34}  \\ 
		SBERT-AFS-large & 53.82 & 53.10 \\ \hline 		
	\end{tabular}
	\caption{Average Pearson correlation $r$ and average Spearman's rank correlation $\rho$ on the Argument Facet Similarity (AFS) corpus \cite{MisraEW16}. Misra et al.\ proposes 10-fold cross-validation. We additionally evaluate in a cross-topic scenario: Methods are trained on two topics, and are evaluated on the third topic.}
	\label{table_afs}
\end{table}

\subsection{Wikipedia Sections Distinction}
\newcite{ein-dor-etal-2018-learning} use Wikipedia to create a thematically fine-grained train, dev and test set for sentence embeddings methods. Wikipedia articles are separated into distinct sections focusing on certain aspects. Dor et al.\ assume that sentences in the same section are thematically closer than sentences in different sections. They use this to create a large dataset of weakly labeled sentence triplets: The anchor and the positive example come from the same section, while the negative example comes from a different section of the same article. For example, from the Alice Arnold article: Anchor: \textit{Arnold joined the BBC Radio Drama Company in 1988.}, positive: \textit{Arnold gained media attention in May 2012.}, negative: \textit{Balding and Arnold are keen amateur golfers.}

We use the dataset from Dor et al. We use the Triplet Objective, train SBERT for one epoch on the about 1.8 Million training triplets and evaluate it on the 222,957 test triplets. Test triplets are from a distinct set of Wikipedia articles. As evaluation metric, we use accuracy: Is the positive example closer to the anchor than the negative example?

Results are presented in Table \ref{table_wikipedia_triplets}. Dor et al.\ fine-tuned a BiLSTM architecture with triplet loss to derive sentence embeddings for this dataset. As the table shows, SBERT clearly outperforms the BiLSTM approach by Dor et al.

\begin{table}[h]
	\centering 
	\footnotesize
	\begin{tabular}{|l|c|}
		\hline
		\textbf{Model} & \textbf{Accuracy}  \\ \hline
		mean-vectors & 0.65 \\
		skip-thoughts-CS & 0.62 \\
		Dor et al.\ & 0.74 \\ \hline 
		SBERT-WikiSec-base & 0.8042 \\ 
		SBERT-WikiSec-large & \textbf{0.8078} \\ 
		SRoBERTa-WikiSec-base & 0.7945 \\
		SRoBERTa-WikiSec-large & 0.7973 \\ \hline
	\end{tabular}
	\caption{Evaluation on the Wikipedia section triplets dataset \cite{ein-dor-etal-2018-learning}. SBERT trained with triplet loss for one epoch.}
	\label{table_wikipedia_triplets}
\end{table}

\section{Evaluation - SentEval} \label{sec_eval_senteval}
\begin{table*}[t]
	\centering 
	\footnotesize
	\begin{tabular}{|l|c|c|c|c|c|c|c||c|}
		\hline
		\textbf{Model} & \textbf{MR} & \textbf{CR} & \textbf{SUBJ} & \textbf{MPQA} & \textbf{SST} & \textbf{TREC} & \textbf{MRPC} & \textbf{Avg.} \\ \hline
		Avg.\ GloVe embeddings & 77.25 & 78.30 & 91.17 & 87.85 & 80.18 & 83.0 & 72.87 & 81.52 \\
		Avg.\ fast-text embeddings & 77.96 & 79.23 & 91.68 & 87.81 & 82.15 & 83.6 & 74.49 & 82.42 \\
		Avg.\ BERT embeddings & 78.66 & 86.25 & 94.37 & 88.66 &  84.40 & 92.8 & 69.45 & 84.94 \\
		BERT CLS-vector & 78.68 & 84.85 & 94.21 & 88.23 & 84.13 & 91.4 & 71.13 &  84.66 \\
		InferSent - GloVe & 81.57 & 86.54 & 92.50 & \textbf{90.38} & 84.18 & 88.2 & 75.77 & 85.59 \\
		Universal Sentence Encoder & 80.09 & 85.19 & 93.98 & 86.70 & 86.38 & \textbf{93.2} & 70.14 & 85.10 \\ \hline
		SBERT-NLI-base & 83.64 & 89.43 & 94.39 & 89.86 & 88.96 & 89.6 & \textbf{76.00} & 87.41 \\
		SBERT-NLI-large & \textbf{84.88} & \textbf{90.07} & \textbf{94.52} & 90.33 & \textbf{90.66} & 87.4 & 75.94 & \textbf{87.69} \\ \hline	
	\end{tabular}
	\caption{Evaluation of SBERT sentence embeddings using the SentEval toolkit. SentEval evaluates sentence embeddings on different sentence classification tasks by training a logistic regression classifier using the sentence embeddings as features. Scores are based on a 10-fold cross-validation.}
	\label{table_senteval}
\end{table*}

SentEval \cite{conneau2018senteval} is a popular toolkit to evaluate the quality of sentence embeddings. Sentence embeddings are used as features for a logistic regression classifier. The logistic regression classifier is trained on various tasks in a 10-fold cross-validation setup and the prediction accuracy is computed for the test-fold. 

The purpose of SBERT sentence embeddings are not to be used for transfer learning for other tasks. Here, we think fine-tuning BERT as described by \newcite{devlin2018bert} for new tasks is the more suitable method, as it updates all layers of the BERT network. However, SentEval can still give an impression on the quality of our sentence embeddings for various tasks.

We compare the SBERT sentence embeddings to other sentence embeddings methods on the following seven SentEval transfer tasks: 
\begin{itemize}
 \item \textbf{MR}: Sentiment prediction for movie reviews snippets on a five start scale \cite{pang-lee-2005-seeing}.

 \item \textbf{CR}: Sentiment prediction of customer product reviews \cite{Hu:2004}.

 \item \textbf{SUBJ}: Subjectivity prediction of sentences from movie reviews and plot summaries \cite{pang-lee-2004-sentimental}.

 \item \textbf{MPQA}: Phrase level opinion polarity classification from newswire \cite{Wiebe2005}.

 \item \textbf{SST}: Stanford Sentiment Treebank with binary labels \cite{socher-etal-2013-recursive}.

 \item \textbf{TREC}: Fine grained question-type classification from TREC \cite{Li2002}.

 \item \textbf{MRPC}:  Microsoft Research Paraphrase Corpus from parallel news sources \cite{Dolan2004}.
\end{itemize}

The results can be found in Table \ref{table_senteval}. SBERT is able to achieve the best performance in 5 out of 7 tasks. The average performance increases by about 2 percentage points compared to InferSent as well as the Universal Sentence Encoder. Even though transfer learning is not the purpose of SBERT, it outperforms other state-of-the-art sentence embeddings methods on this task.

It appears that the sentence embeddings from SBERT capture well sentiment information: We observe large improvements for all sentiment tasks (MR, CR, and SST) from SentEval in comparison to InferSent and Universal Sentence Encoder.

The only dataset where SBERT is significantly worse than Universal Sentence Encoder is the TREC dataset. Universal Sentence Encoder was pre-trained on question-answering data, which appears to be beneficial for the question-type classification task of the TREC dataset.

Average BERT embeddings or using the \texttt{CLS}-token output from a BERT network achieved bad results for various STS tasks (Table \ref{table_sts_tasks}), worse than average GloVe embeddings. However, for Sent\-Eval, average BERT embeddings and the BERT \texttt{CLS}-token output achieves decent results (Table \ref{table_senteval}), outperforming average GloVe embeddings. The reason for this are the different setups. For the STS tasks, we used cosine-similarity to estimate the similarities between sentence embeddings. Cosine-similarity treats all dimensions equally. In contrast, SentEval fits a logistic regression classifier to the sentence embeddings. This allows that certain dimensions can have higher or lower impact on the classification result. 

We conclude that average BERT embeddings / \texttt{CLS}-token output from BERT return sentence embeddings that are infeasible to be used with cosine-similarity or with Manhatten / Euclidean distance. For transfer learning, they yield slightly worse results than InferSent or Universal Sentence Encoder. However, using the described fine-tuning setup with a siamese network structure on NLI datasets yields sentence embeddings that achieve a new state-of-the-art for the SentEval toolkit.

\section{Ablation Study} \label{sec_ablation_study}

We have demonstrated strong empirical results for the quality of SBERT sentence embeddings. In this section, we perform an ablation study of different aspects of SBERT in order to get a better understanding of their relative importance.

We evaluated different pooling strategies (\texttt{MEAN}, \texttt{MAX}, and \texttt{CLS}). For the classification objective function, we evaluate different concatenation methods. For each possible configuration, we train SBERT with 10 different random seeds and average the performances. 

The objective function (classification vs.\ regression) depends on the annotated dataset. For the classification objective function, we train SBERT-base on the SNLI and the Multi-NLI dataset. For the regression objective function, we train on the training set of the STS benchmark dataset. Performances are measured on the development split of the STS benchmark dataset. Results are shown in Table \ref{table_ablation_sbert_classfication}.

\begin{table}[h]
	\centering 
	\footnotesize
	\begin{tabular}{|l|c|c|}
		\hline
		\textbf{} & \textbf{NLI} & \textbf{STSb} \\ \hline
		\multicolumn{3}{|l|}{\textit{Pooling Strategy}} \\ \hline
		\texttt{MEAN} & \textbf{80.78}   & \textbf{87.44} \\
		\texttt{MAX} & 79.07 & 69.92 \\
		\texttt{CLS} & 79.80 & 86.62  \\		
		\hline
		\multicolumn{3}{|l|}{\textit{Concatenation}} \\ \hline
		$(u, v)$ & 66.04 & -\\
		$(|u-v|)$ & 69.78 & - \\
		$(u*v)$ & 70.54 & -\\
		$(|u-v|, u*v)$ & 78.37  & -\\
		$(u, v, u*v)$ & 77.44 & -  \\
		$(u, v, |u-v|)$ &  \textbf{80.78} & - \\
		$(u, v, |u-v|, u*v)$ & 80.44 & - \\ 
		\hline	
	\end{tabular}
	\caption{SBERT trained on NLI data with the classification objective function, on the STS benchmark (STSb) with the regression objective function. Configurations are evaluated on the development set of the STSb using cosine-similarity and Spearman's rank correlation. For the concatenation methods, we only report scores with \texttt{MEAN} pooling strategy. }
	\label{table_ablation_sbert_classfication}
\end{table}

When trained with the classification objective function on NLI data, the pooling strategy has a rather minor impact. The impact of the concatenation mode is much larger. InferSent \cite{conneau2017infersent} and Universal Sentence Encoder \cite{universal_sentence_encoder} both use $(u, v, |u-v|, u*v)$ as input for a softmax classifier. However, in our architecture, adding the element-wise $u*v$ decreased the performance. 

The most important component is the element-wise difference $|u-v|$. Note, that the concatenation mode is only relevant for training the softmax classifier. At inference, when predicting similarities for the STS benchmark dataset, only the sentence embeddings $u$ and $v$ are used in combination with cosine-similarity. The element-wise difference measures the distance between the dimensions of the two sentence embeddings, ensuring that similar pairs are closer and dissimilar pairs are further apart. 

When trained with the regression objective function, we observe that the pooling strategy has a large impact. There, the \texttt{MAX} strategy perform significantly worse than \texttt{MEAN} or \texttt{CLS}-token strategy. This is in contrast to \cite{conneau2017infersent}, who found it beneficial for the BiLSTM-layer of InferSent to use \texttt{MAX} instead of \texttt{MEAN} pooling.

\section{Computational Efficiency} \label{sec_computational_efficiency}

Sentence embeddings need potentially be computed for Millions of sentences, hence, a high computation speed is desired. In this section, we compare SBERT to average GloVe embeddings, InferSent \cite{conneau2017infersent}, and Universal Sentence Encoder \cite{universal_sentence_encoder}.

For our comparison we use the sentences from the STS benchmark \cite{sts2017}. We compute average GloVe embeddings using a simple for-loop with python dictionary lookups and NumPy. InferSent\footnote{\url{https://github.com/facebookresearch/InferSent}} is based on PyTorch. For Universal Sentence Encoder, we use the TensorFlow Hub version\footnote{\url{https://tfhub.dev/google/universal-sentence-encoder-large/3}}, which is based on TensorFlow. SBERT is based on PyTorch. For improved computation of sentence embeddings, we implemented a smart batching strategy: Sentences with similar lengths are grouped together and are only padded to the longest element in a mini-batch. This drastically reduces computational overhead from padding tokens.   

Performances were measured on a server with Intel i7-5820K CPU @ 3.30GHz, Nvidia Tesla V100 GPU, CUDA 9.2 and cuDNN. The results are depicted in Table \ref{table_computational_efficiency}.

\begin{table}[h]
	\centering 
	\footnotesize
	\begin{tabular}{|l|c|c|c|}
		\hline
		\textbf{Model} & \textbf{CPU} & \textbf{GPU}  \\ \hline
		Avg. GloVe embeddings & 6469 & - \\
		InferSent & 137 & 1876 \\ 
		Universal Sentence Encoder & 67 &  1318 \\
		SBERT-base & 44 & 1378 \\
		SBERT-base - smart batching & 83 & 2042 \\
		\hline			
	\end{tabular}
	\caption{Computation speed (sentences per second) of sentence embedding methods. Higher is better.}
	\label{table_computational_efficiency}
\end{table}

On CPU, InferSent is about 65\% faster than SBERT. This is due to the much simpler network architecture. InferSent uses a single Bi\-LSTM layer, while BERT uses 12 stacked transformer layers. However, an advantage of transformer networks is the computational efficiency on GPUs. There, SBERT with smart batching is about 9\% faster than InferSent and about  55\% faster than Universal Sentence Encoder. Smart batching achieves a speed-up of 89\% on CPU and 48\% on GPU. Average GloVe embeddings is obviously by a large margin the fastest method to compute sentence embeddings. 

\section{Conclusion}
We showed that BERT out-of-the-box maps sentences to a vector space that is rather unsuitable to be used with common similarity measures like cosine-similarity. The performance for seven STS tasks was below the performance of average GloVe embeddings. 

To overcome this shortcoming, we presented Sentence-BERT (SBERT). SBERT fine-tunes BERT in a siamese / triplet network architecture. We evaluated the quality on various common benchmarks, where it could achieve a significant improvement over state-of-the-art sentence embeddings methods. Replacing BERT with RoBERTa did not yield a significant improvement in our experiments.

SBERT is computationally efficient. On a GPU, it is about 9\% faster than InferSent and about 55\% faster than Universal Sentence Encoder. SBERT can be used for tasks which are computationally not feasible to be modeled with BERT. For example, clustering of 10,000 sentences with hierarchical clustering requires with BERT about 65 hours, as around 50 Million sentence combinations must be computed. With SBERT, we were able to reduce the effort to about 5 seconds.

\section*{Acknowledgments}
This work has been supported by the German Research Foundation through the German-Israeli Project Cooperation (DIP, grant DA 1600/1-1 and grant GU 798/17-1). It has been co-funded by the German Federal Ministry of Education and Research (BMBF) under the promotional references 03VP02540 (ArgumenText).

\bibliography{emnlp-ijcnlp-2019}
\bibliographystyle{acl_natbib}

\end{document}